%% file: iclr2026_conference.tex
\documentclass{article} 
\usepackage{iclr2026_conference,times}

\input{math_commands.tex}

\usepackage{hyperref}
\usepackage{graphicx}
\usepackage{url}
\usepackage{subcaption}
\usepackage{booktabs}
\usepackage{svg} 
\usepackage{enumitem}
\usepackage{mathtools}

\newcommand{\defeq}{\coloneqq}

\title{Multi-state Protein Design with \\ DynamicMPNN}

\author{
Alex Abrudan\thanks{Equal contribution.}, \\
Yusuf Hamied Department of Chemistry \\
University of Cambridge \\
Cambridge, UK \\
\texttt{\{aca41\}@cam.ac.uk}
\And 
Sebastian Pujalte Ojeda\footnotemark[1], \\
Yusuf Hamied Department of Chemistry \\
University of Cambridge \\
Cambridge, UK \\
\texttt{\{sp2120\}@cam.ac.uk}
\And
Chaitanya K. Joshi \\
Department of Computer Science and Technology \\
University of Cambridge \\
Cambridge, UK \\
\And
Matthew Greenig, \\
Yusuf Hamied Department of Chemistry \\
University of Cambridge \\
Cambridge, UK \\
\And
Felipe Engelberger \\
Institute for Drug Discovery \\
Leipzig University \\
Leipzig, Germany \\
\And
Alena Khmelinskaia \\
Department of Chemistry \\
Ludwig-Maximilans-University \\
Munich, Germany \\
\And
Jens Meiler \\
Institute for Drug Discovery \\
Leipzig University \\
Leipzig, Germany \\
\And
Michele Vendruscolo \\
Yusuf Hamied Department of Chemistry \\
University of Cambridge \\
Cambridge, UK \\
\And
Tuomas P. J. Knowles \\
Yusuf Hamied Department of Chemistry \\
University of Cambridge \\
Cambridge, UK \\
}

\iclrfinalcopy 
\begin{document}

\maketitle

\begin{abstract}
Structural biology has long been dominated by the \emph{one sequence, one structure, one function} paradigm, yet many critical biological processes—from enzyme catalysis to membrane transport—depend on proteins that adopt multiple conformational states. Existing multi-state design approaches rely on post-hoc aggregation of single-state predictions, achieving poor experimental success rates compared to single-state design. We introduce DynamicMPNN, an inverse folding model explicitly trained to generate sequences compatible with multiple conformations through joint learning across conformational ensembles. Trained on 46,033 conformational pairs covering 75\% of CATH superfamilies and evaluated using Alphafold 3, DynamicMPNN outperforms ProteinMPNN by up to 25\% on decoy-normalized RMSD and by 12\% on sequence recovery across our challenging multi-state protein benchmark.
\end{abstract}

\section{Introduction}
A commonly derived assumption from Anfinsen’s experiment is that proteins
adopt only one native 3D structure, leading to the ``one sequence, one structure, one function'' canon.
This view has been indirectly reinforced by the predominant use of X-Ray crystallography in experimental protein structure determination, which requires that proteins form a homogenous, diffractible crystal to be characterised \citep{dishman2018}. The large collection of static protein structures in the PDB  has enabled the development of high-accuracy machine learning models for tasks such as structure prediction \citep{Jumper2021,casp} and inverse folding \citep{proteinmpnn, hsu2022}. Amongst contemporary inverse folding models, ProteinMPNN has been particularly widely adopted in applied protein design projects due to its low inference costs and robust experimental success rates \citep{proteinmpnn, rfdiffusion, solublempnn}, outperforming traditional physics-based design methods on both fronts \citep{rosettadesign}.

\begin{figure*}[t!]
\begin{center}
\centerline{\includegraphics[width=\textwidth]{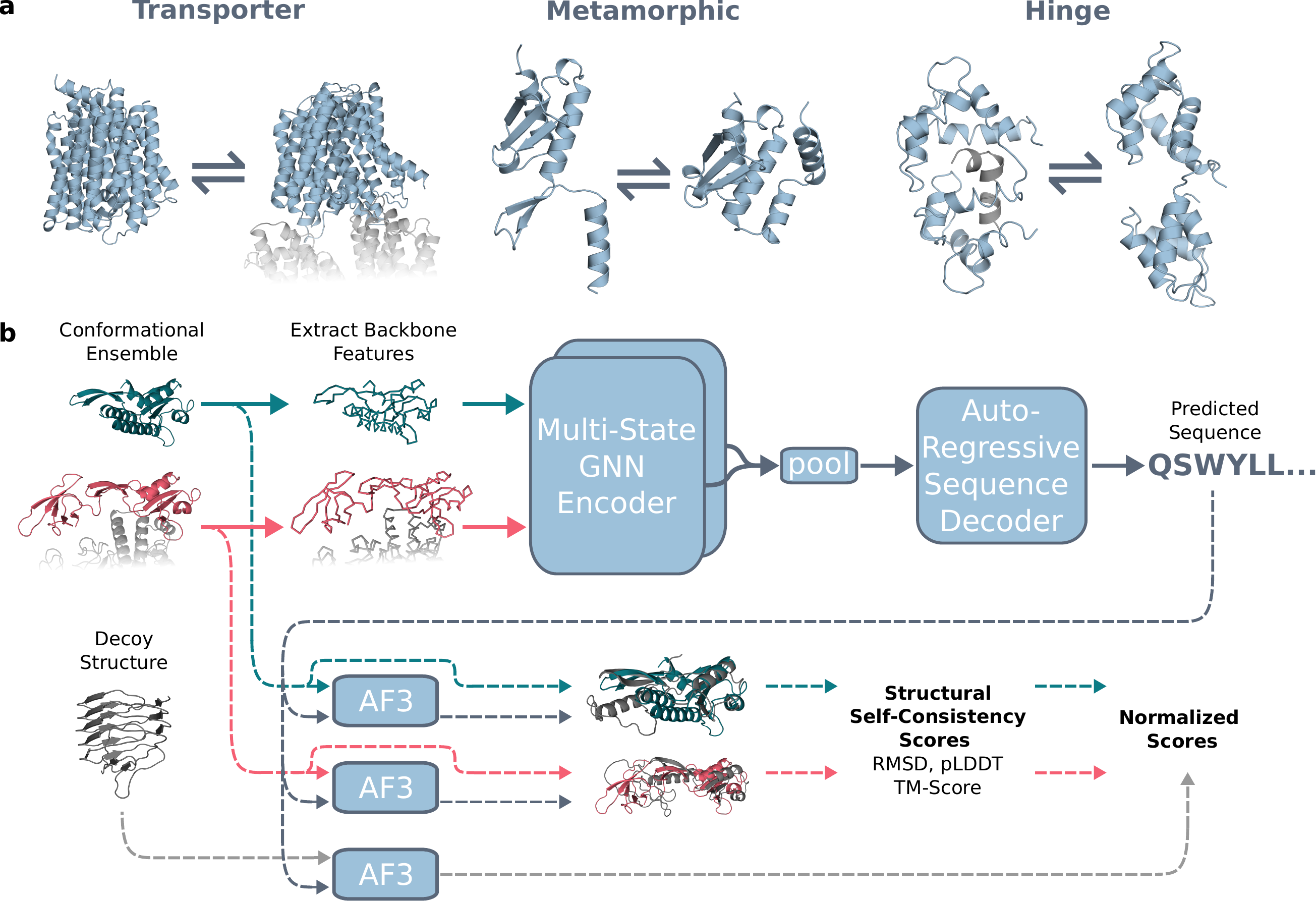}}
\caption{
DynamicMPNN for multi-state protein design.
(a) Examples of proteins with multiple conformational states: transporters in closed and open states (PDB: 6NC7, 6NC9), metamorphic protein with alternative folds (PDB: 4QHH, 4QHF) and hinges showing domain movement (PDB: 5D0W, 1CFC). 
(b) Schematic of DynamicMPNN, an inverse folding model trained to generate protein sequences with multiple conformational states. Conformations are encoded with their respective chemical environments (i.e. interaction partners shown in gray). Solid lines show the flow of information in the model, while dashed lines show the evaluation pipeline using AlphaFold 3 (AF3); employing target structures as templates during inference and measuring the deviations between predicted and target structures, with decoy structures serving as negative controls.}
\label{fig1}
\end{center}
\end{figure*}

Although direct experimental characterisation of protein dynamics remains a challenge, the conformational diversity of proteins underlies crucial biological functions such as enzyme catalysis, protein-protein interactions, allostery, and human disease \citep{codnas}. In applied protein design, bio-switches - proteins that switch between two structural states - are of particular importance, with key applications in engineering artificial bio-motors, signalling pathways, biosensors, or drug delivery systems \citep{STEIN2015101, Praetorius2023}.
While most known switches undergo rearrangements in the context of a single fold \citep{AMBROGGIO2006525}, the class of metamorphic proteins undergo changes in both their secondary structure and fold (Fig. \ref{fig1}a)
and have been predicted to represent up to 4\% of the PDB chains \citep{extandfoldswitching}.
These proteins typically adopt two main functional states \citep{dishman2018} and a finite number of conformations (see Discussion). Beyond the world of switches, other dynamic proteins are characterised by continuous conformational landscapes (e.g. intrinsically disordered proteins \citep{fuzzy}).

Multi-state protein design was first achieved through rational design and physics-based methods such as RosettaDesign \citep{rosettadesign}. Previous campaigns leveraging these methods have attempted to design metamorphic metal-binding peptides \citep{Ambroggio2006,Zico}, closely related sequences that adopt diverging folds \citep{Wei2020}, and hinge proteins with binder-regulated thermodynamic equilibria, allowing the relative populations of different structural states to be modulated by exogenous proteins \citep{Zhang2022,Rubio2021, Praetorius2023}. More recently, \citet{Praetorius2023} developed ProteinMPNN Multi-state Design (ProteinMPNN-MSD) \citep{proteinmpnn}: an inference strategy for extending ProteinMPNN inverse folding to multiple structural states by averaging the logits of two independent single-state ProteinMPNN embeddings during the decoding step. In a similar vein, ProteinGenerator \citep{proteingenerator} applies a sequence diffusion process conditioned on structural predictions by RosettaFold \citep{rosettafold}, and attempts to design sequences that adopt multiple states by averaging the sequence logits predicted by the model across distinct target structures.

Still, significant limitations in current multi-state design pipelines remain. Designs generated from ProteinMPNN-MSD faced a rigorous selection process. From an initial pool of over two million computational designs, subsequent filtering and experimental validation yielded 46 soluble 2-state hinge sequences. Among these, nine successfully demonstrated binding to the target peptide \citep{Praetorius2023}. Likewise, the authors of ProteinGenerator reported a significantly lower \textit{in silico} design success rate of 0.05\% for their multi-state design task compared to rates of 2-10\% observed for various single-state sequence design objectives using their framework \citep{proteingenerator}. These observations - combined with the relative scarcity of published data on ML-driven multi-state \textit{de novo} design campaigns - suggest that current ML methods for multi-state protein design have been significantly less successful than their single-state counterparts. We propose that this gap can be attributed to limited multi-conformational datasets, weak benchmarks, and the poor performance of folding models in predicting alternative states \citep{Chakravarty2024} (which adversely affects their efficacy as self-consistency filters in protein design workflows).

\textbf{Our contributions.}
This paper introduces DynamicMPNN (Fig. \ref{fig1}b), a novel geometric deep learning-based pipeline for multi-state protein sequence design.
\begin{itemize}[itemsep=0.3em, topsep=0.3em, parsep=0em, partopsep=0em]
    \item DynamicMPNN is the first explicit multi-state inverse folding model for protein design. To train DynamicMPNN, we create a new ML-ready dataset of proteins with multiple conformations using the PDB and CoDNaS \citep{codnas} databases, and evaluate the method on 96 biologically relevant metamorphic, hinge, and transporter proteins.

    \item We introduce a novel data processing pipeline and architecture able to handle sequence-aligned structure pairs with heterogeneous sequences, enabling us to leverage the conformational diversity across non-identical protein states.
    
    \item We propose a multi-state self-consistency metric and benchmark based on \emph{Alphafold3} (AF3) \citep{abramson2024accurate} using target structures as templates.
    
    \item  DynamicMPNN improves performance over ProteinMPNN \citep{proteinmpnn} on AF3 by up to 25\% on RMSD and 8\% on pLDDT decoy normalized self-consistency values.
    
\end{itemize}

\section{The DynamicMPNN pipeline}

\subsection{Protein multi-conformational dataset}
\label{sec:dataset}

\begin{figure}[t!]
\begin{center}
\centerline{\includegraphics[width=0.75\textwidth]{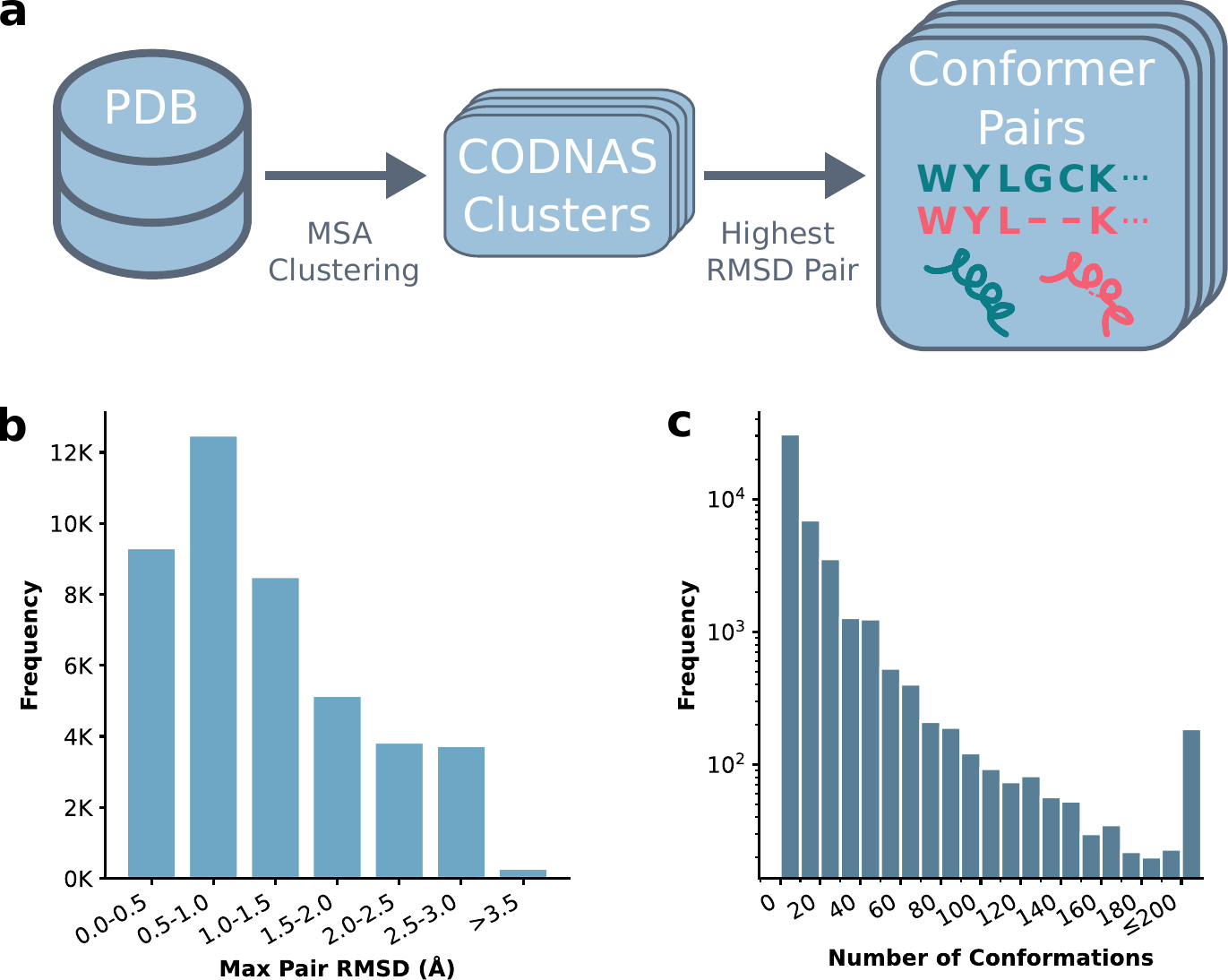}}
\caption{
Multi-state protein dataset.
(a) Data processing pipeline used to construct sequence-aligned structure pairs.  (b) Distribution of the number of conformations per CoDNaS cluster. (c) Distribution of the maximum C$\alpha$-RMSD between pairs of structures in each CoDNaS cluster.
}
\label{fig2}
\end{center}
\vspace{-5mm}
\end{figure}

While over 900,000 individual protein chains (sequence-structure pairs) are available in the PDB, multi-conformational data is at a first glance far more scarce with only roughly 12,000 NMR-derived protein ensembles covering just 21\% of CATH superfamilies. To overcome this limitation, we exploit the sequence redundancy in the PDB: following \citet{codnas} and \citet{pdbflex}, we cluster all protein chain sequences with a very high sequence similarity threshold ($\geq$95\%) to build a multi-conformational dataset of 46,033 conformational clusters with at least 2 members - expanding coverage to 75\% of CATH superfamilies (Fig. \ref{fig2}a).
The 95\% similarity threshold accounts for soluble tags or point mutations/alterations performed in different experiments which are unlikely to affect the overall fold. Clusters contain highly unbalanced conformational populations (Fig. \ref{fig2}c): a relatively small number of clusters contain thousands of sequences, while the majority contain only a few. This contributes to the relatively small intra-cluster RMSD in the majority of the clusters (Fig. \ref{fig2}b), since many alternative protein states live in low populations.

To maximise the conformational signal and minimize alignment artifacts, we built our dataset using only pairs of chains from one or two PDB entries that have the largest RMSD from each cluster (Fig. \ref{fig2}c). 
Critically, this selection emphasizes our focus on proteins exploring a finite number of functional states, with most switching between just two conformations \citep{dishman2018,leaverfay2011generic,alberstein2022design}. While we hope to explore design of more continuous conformational spaces in future work, we focus presently on training DynamicMPNN to design two-state proteins.
Furthermore, molecular dynamics datasets were opted against due to the poor training signal given by limited sequence variability and scarcity of trajectories showing large conformational changes \citep{atlasMD,mdcath}, which normally happen across long biological timescales. At the other end of the spectrum, single-chain disordered protein simulations \citep{tesei2024conformational} expose continuous conformational landscapes which are hard to discretize and have a high degree of stochasticity with a low signal-to-noise ratio rather than the ordered topologies of globular multi-state proteins \citep{fuzzy}.

For dataset splitting, we first curate a benchmark composed of five previous studies of proteins with large 2-state conformational changes: (1) 92 metamorphic proteins \citep{extandfoldswitching}, (2) 91 apo-holo proteins \citep{AF-apoholo}, (3) the OC23 and OC85 open-closed datasets \citep{Kalakoti2025}, and (4) 20 transporter proteins \citep{Kalakoti2025}.
The proteins with the highest inter-state RMSD were assigned to the test set (96 samples), while the rest were assigned to the validation set (100 samples). Training clusters were filtered to exclude any with TM-score $> 0.4$ \citep{zhang2004scoring} and $>30\%$ sequence similarity to test/validation structures, preventing structural similarity leakage and yielding a final training set of 44,243 conformer pairs.

We additionally curate a set of single-state sequence-structure pairs from the 30\% sequence similarity clusters not represented in our multi-conformational training dataset ($n=27,394$). This augmentation strategy maximizes coverage of protein fold space while preserving the multi-conformational learning signal.

See Appendix \ref{app:data} for further details on dataset composition.

\subsection{DynamicMPNN for multi-state inverse folding}

Single-state inverse folding methods seek to model the conditional distribution $p(Y|X)$ where $X \in \mathbb{R}^{n \times 3 \times 3}$ represents a protein backbone with $n$ residues, and $Y = (y_1, \ldots, y_n)$ is the amino acid sequence. Extensions of these methods to multi-state design have thus far been limited to post-hoc aggregation of independent single-state predictions. We believe such methods favour logits which are highly biased towards one conformation, whose average over the 2 states is higher than the one of logits which are moderately valued across both states \citep{grnade}.
Instead, DynamicMPNN learns the joint conditional distribution of $p(Y|X_1, \ldots, X_m)$ directly through autoregressive sequence generation, where $\{X_1, \ldots, X_m\}$ represent distinct protein conformations encoded into a shared latent space; thus learning a sequence distribution that simultaneously satisfies multiple structural constraints.
We decompose this joint conditional probability using the autoregressive factorization:

\begin{gather}
    \mathit{p(Y|X_1,...,X_m)} = \nonumber\\
    \mathit{\prod_{i=1}^n p(y_i|y_{i-1},...,y_1;X_1,...,X_m)}
\end{gather}

where each factor represents the probability of selecting residue $y_i$ given the sequence prefix and the complete structural ensemble.

\textbf{Overall architecture}. DynamicMPNN independently encodes each of the functional states of a protein, together with their binding partners, into a shared latent feature space (Fig. \ref{fig1}b). Embeddings of the chains-to-be-designed chains are then pooled across conformations to obtain a single embedding from which a sequence is auto-regressively generated.

Our architecture is built upon gRNAde \citep{grnade}, a multi-state GNN for RNA inverse folding. 
For both encoder and decoder, we employ SE(3)-equivariant Geometric Vector Perceptron \citep{jing2021} layers which maintain computational efficiency through edge sparsity (k-NN edges with k=32). 
Within the GVP, the scalar features $\vs_i \in \mathbb{R}^{k \times f}$ and vector features $\vec{\vv}_i \in \mathbb{R}^{k \times f' \times 3}$ for each node $i$ \citep{duval2024hitchhikersguidegeometricgnns}:
\begin{align}
    \vm_i, \vec{\vm}_i     & \defeq \sum_{j \in \mathcal{N}_i} \textsc{Msg} \big( \left(\vs_i, \vec{\vv}_i\right) , \left(\vs_j, \vec{\vv}_j\right) , \ve_{ij} \big) , \\
    \vs_i', \vec{\vv}_i' & \defeq \textsc{Upd} \big( \left(\vs_i, \vec{\vv}_i\right) \ , \ \left(\vm_i, \vec{\vm}_i\right) \big) ,
\end{align}
where $\textsc{Msg}, \textsc{Upd}$ are Geometric Vector Perceptrons, a generalization of MLPs to take tuples of scalar and vector features as input and apply $O(3)$-equivariant non-linear updates.
The overall GNN encoder is $SO(3)$-equivariant due to the use of reflection-sensitive input features (dihedral angles) combined with $O(3)$-equivariant GVP-GNN layers. Both encoder and decoder are assigned 8 GVP layers, following findings in \citet{hsu2022} (2022). 
See Appendix \ref{app:model} for further details.

We present two different encoder architectures: 
\begin{itemize}
    \item \textbf{DynamicMPNN}: Independent encoder channels for each conformation, followed by Deep Set pooling \citep{zaheer2017deep} - it is invariant to conformation order and does not add extra parameters to our model. We note that while some more expressive pooling strategies have been shown to provide marginal performance improvements, they usually come at a great cost in efficiency \citep{grnade}. Only node features are updated. 
    \item \textbf{DynamicMPNN + DSS}: We implement cross-attention between the encoder channels after each layer using a Deep Symmetric Set (DSS) \citep{dss} module, which allows for richer inter-conformation interactions at the cost of increased computational complexity. In the scatter/gather DSS strategy, node embeddings of all design chains are averaged, passed through GVP layers, and added back to the residual features of each channel. Both node and edge features are updated.  
\end{itemize}

A major distinction from gRNAde is that DynamicMPNN models multi-chain complexes and can thus condition conformational changes on protein binders and oligomeric states. It may be possible to use this principle to engineer controllable protein conformational switches by tuning the free energy difference between folds and their binding interactions \citep{alberstein2022design}.
However, this also introduces challenges in dealing with nonidentical sequences and missing residues across conformations, which we address as follows.

\begin{figure*}[t!]
\begin{center}
\centerline{\includegraphics[width=0.9\textwidth]{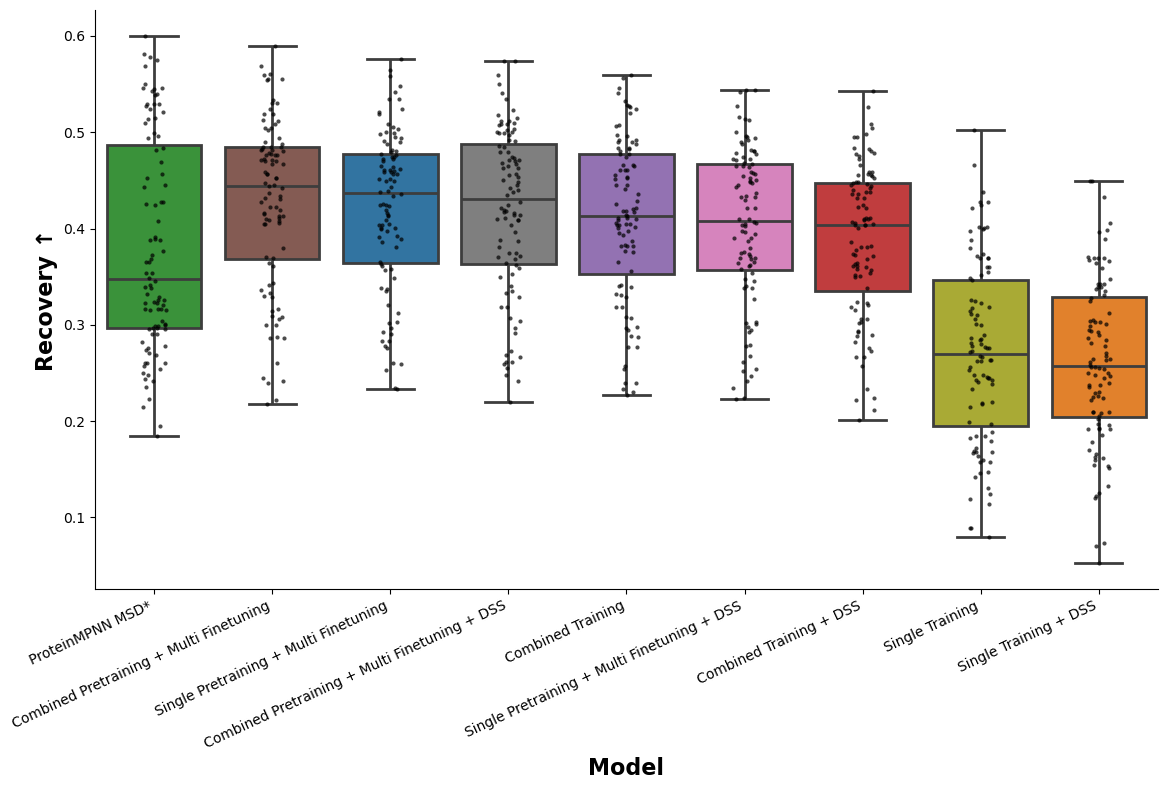}}
\caption{Sequence recovery performance across DynamicMPNN model variants and ProteinMPNN baseline on multi-state protein benchmark ($n=96$). Combined training approaches achieve highest performance, with models that only incorporate multi or single state training data performing poorly.}
\label{fig:seq_recov}
\end{center}
\end{figure*}

\textbf{Alignment and pooling.}
To deal with nonidentical sequences and missing residues during training, structure pairs are featurized and encoded independently, and subsequently aggregated based on pairwise sequence alignments (i.e. only nongap residues are taken into account when pooling).

\textbf{Multi-chain encoding and masking}.
Given that conformational shifts often depend on interactions, DynamicMPNN encodes the chemical environment for each conformational state (currently limited to proteins). During training, we expose the sequence information of binding partners in the encoding, only masking chains with $>$70\% sequence similarity to the ground truth sequence of the chains of interest.

\subsection{Multi-state design evaluation}
Following previous work \cite{wang2023pdb}, we evaluate the refoldability of generated sequences rather than sequence recovery (Appendix \ref{app:supp}). Existing refoldability methods compare target 
structures to single conformations predicted by folding models (e.g., AlphaFold2 \cite{Jumper2021}). We argue that this approach is unsuitable for multi-state design since folding models typically predict one dominant state or interpolate between conformations rather than sampling the full conformational ensemble \cite{lane2023protein,Chakravarty2024,AF-apoholo}. Furthermore, as mentioned in Section~\ref{sec:dataset}, recent DL-based protein folding models have been shown not to learn protein folding physics, but rather to compress the protein space information while preserving lower complexity motifs such as sequence segment pairing \citep{outeiral2022current,llmmotifs,Chakravarty2024}.

We propose using a template-based Alphafold 3 (AF3) framework \cite{abramson2024accurate}: for each de novo sequence, we perform 2 AF3 runs, one with each conformational state given as template to moderately bias the model towards the target conformation. The similarity - C$\alpha$-RMSD or TM-score \citep{zhang2004scoring} - between predicted structures and the templates, along with AF3 confidence scores, 
serve as a proxy for the likelihood that the designed sequence will fold into the target structure. This framework naturally enables evaluation of refoldability across multiple conformations.

Formally, for a protein with conformational states $X = \{X_1, X_2, \dots, X_m\}$ and designed sequence $Y$, we define the AF3 RMSD for each target conformation $X_k$ as:

\begin{equation}
    \text{AF3}_{template}(Y, X_k) = \text{RMSD}(\text{AF3}(Y, X_k), X_k)
\end{equation}
where $\text{AF3}_{template}(Y, X_k)$ is the structure predicted by AlphaFold3 for sequence $Y$ when a template of $X_k$ is provided. To account for the structural bias induced by the template, we define a normalization strategy to contextualize observed deviations:

\textbf{Decoy normalization (Decoy Norm)}: We provide AF3 structurally dissimilar decoy structures as templates ($\text{TM-score}< 0.4$) using the same sequences designed and measure the resulting deviations. This control assesses whether sequences fold specifically into their targets or may fold equally well into arbitrary structures:
\begin{equation}
    \text{RMSD}_{\text{decoy}}(Y, X_k; D) = \frac{\text{AF3}_{template}(Y, X_k)}{\text{AF3}_{template}(Y, D)}
\end{equation}
where $D$ is a decoy that is structurally dissimilar to $X_k$.
Additionally, we measure pLDDT confidence scores to evaluate AF3 fold uncertainty. High RMSD with low pLDDT indicates poor template matching, while low RMSD with high pLDDT suggests a successful design.

\begin{table}[ht!]
\caption{Refoldability performance comparison of DynamicMPNN model variants on multi-state protein design benchmark ($n=96$). Raw metrics show absolute performance values, while normalized metrics show performance relative to random decoy structures.}
\label{tab:model_performance}
\centering
\setlength{\tabcolsep}{2.5pt}
\resizebox{\linewidth}{!}{
\begin{tabular}{l*{3}{c}|*{3}{c}}
\toprule
& \multicolumn{3}{c|}{\textbf{Raw Metrics}} & \multicolumn{3}{c}{\textbf{Decoy-Normalized Metrics}} \\
\cmidrule(lr){2-4} \cmidrule(lr){5-7}
\textbf{Model Variant} & \textbf{pLDDT $\uparrow$} & \textbf{RMSD (Å) $\downarrow$} & \textbf{TM-score $\uparrow$} & \textbf{pLDDT $\uparrow$} & \textbf{RMSD $\downarrow$} & \textbf{TM-score $\uparrow$} \\
\midrule
Combined Training & \textbf{79.28} (9.28) & \textbf{4.37} (4.76) & \textbf{0.774} (0.246) & 1.513 (0.503) & \textbf{0.200} (0.183) & 7.396 (2.192) \\
Combined Training + DSS & 78.87 (8.82) & 4.38 (4.47) & 0.769 (0.239) & \textbf{1.532} (0.524) & \textbf{0.200} (0.164) & \textbf{7.384} (2.304) \\
Single Training & 62.59 (16.84) & 13.48 (13.02) & 0.503 (0.350) & 1.466 (0.537) & 0.479 (0.339) & 5.014 (3.068) \\
\bottomrule
\end{tabular}
}
\label{tab:rmsd_mpnn}
\begin{center}
\vspace{-3mm}
\end{center}
\end{table}

\section{Results and Discussion}

\begin{table}[ht!]
\vspace{-2mm}
\caption{Sequence recovery performance comparison across DynamicMPNN model variants and ProteinMPNN baseline on multi-state protein design benchmark.}
\centering
\setlength{\tabcolsep}{3pt}
\resizebox{.75\linewidth}{!}{
\begin{tabular}{lc}
\toprule
\textbf{Model Variant} & \textbf{Sequence Recovery (\%) $\uparrow$} \\
\midrule
Combined Pretraining + Multi Finetuning & \textbf{42.7} (8.8) \\
Single Pretraining + Multi Finetuning & 42.1 (8.3) \\
Combined Pretraining + Multi Finetuning + DSS & 41.9 (8.6) \\
Combined Training & 41.0 (8.5) \\
Single Pretraining + Multi Finetuning + DSS & 40.3 (8.2) \\
Combined Training + DSS & 38.8 (7.8) \\
ProteinMPNN MSD* & 38.0 (11.0) \\
Single Training & 27.1 (9.4) \\
Single Training + DSS & 26.2 (8.5) \\
\bottomrule
\end{tabular}
}
\label{tab:seq_recov}
\begin{center}
\end{center}
\vspace{-2mm}
\end{table}

\begin{table}[ht!]
\caption{Performance comparison of DynamicMPNN variants and ProteinMPNN baseline on subset ($n=61$) of multi-state protein design benchmark. Standard deviations shown in parentheses. Note that ProteinMPNN MSD's handling of gap tokens and missing residues (i.e., X tokens) limited the number of designs that could be refolded using AF3, necessitating separate comparison.}
\label{tab:mpnn_comparison}
\centering
\setlength{\tabcolsep}{2.5pt}
\resizebox{\linewidth}{!}{
\begin{tabular}{l*{3}{c}|*{3}{c}}
\toprule
& \multicolumn{3}{c|}{\textbf{Raw Metrics}} & \multicolumn{3}{c}{\textbf{Decoy-Normalized Metrics}} \\
\cmidrule(lr){2-4} \cmidrule(lr){5-7}
\textbf{Model} & \textbf{pLDDT $\uparrow$} & \textbf{RMSD (Å) $\downarrow$} & \textbf{TM-score $\uparrow$} & \textbf{pLDDT $\uparrow$} & \textbf{RMSD $\downarrow$} & \textbf{TM-score $\uparrow$} \\
\midrule
Combined Training & \textbf{79.23} (7.79) & \textbf{4.30} (3.50) & \textbf{0.748} (0.233) & 1.401 (0.411) & \textbf{0.206} (0.170) & \textbf{7.093} (2.034) \\
Combined Training + DSS & 78.90 (7.68) & 4.56 (3.52) & 0.737 (0.225) & 1.404 (0.425) & 0.217 (0.170) & 7.090 (2.384) \\
Single Training & 63.10 (15.65) & 13.21 (12.18) & 0.477 (0.331) & 1.364 (0.458) & 0.508 (0.337) & 4.527 (2.777) \\
ProteinMPNN MSD* & 74.29 (11.37) & 5.84 (4.56) & 0.689 (0.244) & \textbf{1.416} (0.389) & 0.275 (0.219) & 6.533 (2.325) \\
\bottomrule
\end{tabular}
}
\begin{center}
\label{tab:rmsd_mpnn_2}
\end{center}
\vspace{-3mm}
\end{table}

\textbf{Setup}. We evaluate several training strategies and model architectures to assess the impact on multi-state learning. Single Training models follow the standard single state inverse folding approach, learning exclusively from single-state sequence structure pairs, similarly to ProteinMPNN \cite{proteinmpnn}. In Combined Training, models are presented with both single and multi-state training samples. We also explore training over multiple stages, where single-state data is provided during pretraining to teach general principles of protein geometry from abundant single-state structures before specializing on multi-state conformational pairs (Multi Finetuning). For each model, the training checkpoint with the highest sequence recovery on the multi-state validation set was selected for evaluation.

All DynamicMPNN versions were trained on either 8 A100-80GB or 8 H100-80GB GPUs using a batch size of 32 and Adam \citep{kingma2014adam} optimizer with learning rate $10^{-3}$. Training for each stage was run until the respective validation sequence recovery (on single-state validation dataset for single training runs and multi-state for multi-state runs) plateaued or began to decline, typically after 20-50 epochs. Then, DynamicMPNN and ProteinMPNN (using Multi-state Design inference strategy) were used to sample 16 sequences for the benchmark test set of 96 paired conformations, which were each run through the AF3 pipeline separately against both target states.

Refoldability was evaluated for designed sequences of the model checkpoints with the highest sequence recovery scores by predicting the structure using AF3 with target and decoy structures as templates. Aggregated evaluation metrics were computed by averaging across all 16 sequences and both states. 

\textbf{DynamicMPNN outperforms existing benchmarks across multiple evaluation metrics.} Our best-performing model achieves substantial improvements over baseline methods: a 25\% reduction in decoy-normalized RMSD (Tab. \ref{tab:rmsd_mpnn_2}) and a 12\% improvement in sequence recovery  (Fig.\ref{fig:seq_recov}; Tab. \ref{tab:seq_recov}) compared to ProteinMPNN Multi-State Design (MSD). This performance gain is particularly noteworthy considering that ProteinMPNN's training dataset contains proteins within 30\% sequence similarity clusters of 91 out of 96 benchmark proteins, making our improvement on this established baseline significant. We leave the retraining of ProteinMPNN on a rigorous train-test split for future work.

However, a more rigorous comparison involves our single-state trained model, which adheres to the same training-test split and thus eliminates potential data leakage concerns. Against this baseline, DynamicMPNN demonstrates substantial improvements across all evaluation metrics, providing clear evidence that explicit multi-state training is superior to single-state approaches for multi-conformational protein design tasks (Tab. \ref{tab:seq_recov}). Our single-state trained model performs considerably worse than both DynamicMPNN variants and ProteinMPNN MSD, achieving only 63.10 pLDDT compared to 79.23 for our best model, and showing nearly three-fold higher RMSD values (13.21 Å vs 4.30 Å; Tab. \ref{tab:model_performance}). Some likely reasons to explain the under performance of our model checkpoint trained exclusively on single-state data is the aforementioned unfair train splitting of ProteinMPNN. 

\textbf{Combined and multi-state finetuning training strategies far outperform the single-state only training} 
Models trained only on single-state data (Single Training) perform poorly on multi-state design as they are unable to create and decode a meaningful latent representation of the conformational changes. The Combined Training approach, which exposes models to both single-state and multi-state examples during training, achieves the optimal balance and consistently outperforms other training strategies.



While DynamicMPNN AF3 RMSD values are still high, it is important to contextualize these results within our deliberately challenging benchmark: our test set comprises proteins with the largest documented conformational changes in the known proteome. This can be seen as an extrapolation beyond the overall modest conformational changes observed in most PDB structures and hence our training data.

\section{Conclusion}
We present DynamicMPNN, the first explicit multi-state inverse folding model, achieving up to 25\% improvement over ProteinMPNN on our multi-state benchmark. 
By jointly learning across conformational ensembles rather than aggregating single-state predictions, DynamicMPNN captures sequence constraints required for multiple functional conformations. This opens possibilities for engineering synthetic bioswitches, allosteric regulators, and molecular machines.
It is unclear if the presented one-size-fits-all approach to multi-state design will be effective experimentally, or if specialized models for different classes of conformational changes depending on their thermodynamic complexity will be beneficial.






\bibliography{iclr2026_conference}
\bibliographystyle{iclr2026_conference}

\appendix
\section{Appendix}

\subsection{LLM acknowledgement}
The authors acknowledge that they have used LLMs in the process of paper writing for style suggestions and proofreading.

\section{Supplementary Results}
\label{app:supp}

\subsection{DynamicMPNN preliminary designs}

\begin{figure}[ht]
\begin{center}
\centerline{\includegraphics[width=\columnwidth]{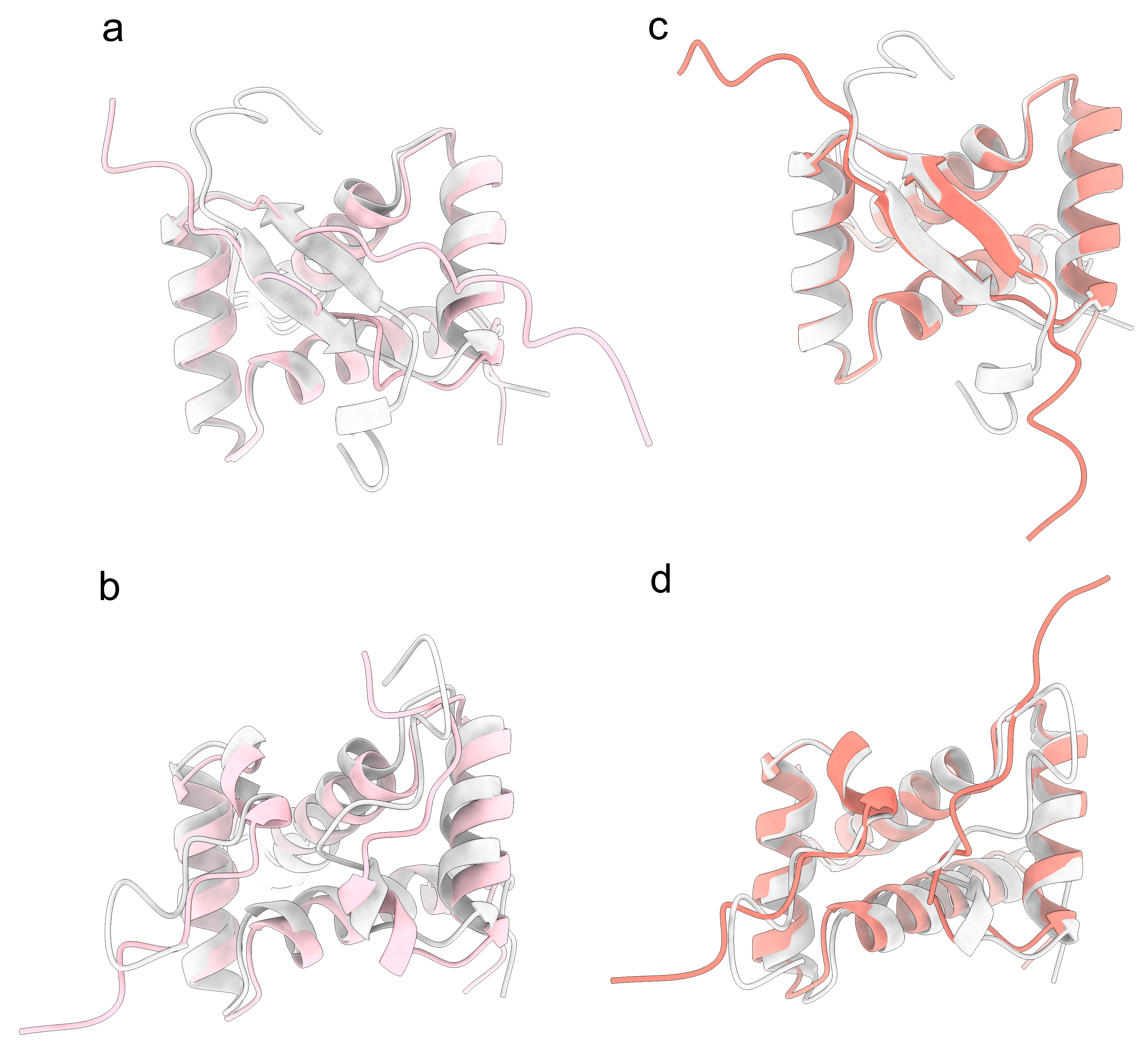}}
\caption{
Switch Arc protein case study.
(a, b) ProteinMPNN  and (c,d) DynamicMPNN best design structure prediction (pink and salmon, respectively) against both Arc states from PDB ID: 1BDT and 1QTG respectively (grey). The DynamicMPNN design recapitulates the beta sheet fold (c), but the ProteinMPNN design does not (a).  
}
\label{fig3}
\end{center}
\vspace{-5mm}
\end{figure}


\section{Dataset Details}
\label{app:data}
To construct the dataset, we obtained 46,033 Multiple Sequence Alignment (MSA) clusters at $\geq95\%$ local sequence similarity from the latest version of CoDNaS (\texttt{v2025}) \citep{codnas}, including NMR model structures. Importantly, the CoDNaS dataset prevents potential errors created by mixing different homologues in the same cluster by enforcing the same UniProt ID for all custer members. All available conformations of a protein are included - as different experimental conditions and sequence variations can reveal distinct thermodynamic states of the same protein \citep{michele_pdb_redundancy}. While clusters contain varying numbers of conformations (Fig. \ref{fig2}b) we constructed our dataset using only pairs of chains from one or two PDB entries that have the largest RMSD from each cluster (Fig. \ref{fig2}c). Chosen pairs represent the most distinct conformational states. We will include conformational information of the whole cluster in future work.

While other inverse folding models \citep{hsu2022} saw improved performance when trained on orders of magnitude of more protein structures from AFDB, previous studies have found that the majority of high-confidence structures in AFDB map to known CATH superfamilies \citep{bordin2023alphafold2}, and that AlphaFold struggles in predicting alternative states \citep{Chakravarty2024}. We therefore decided against including AFDB structures in our training set.

\section{Model Details}
\label{app:model}

\subsection{Featurisation scheme}

We use a similar featurisation scheme as in \citep{jamasb2024evaluating}. 
\textbf{Node scalar features} are transformer-like positional encoding in a 16-dimensional array; backbone dihedral angles $\mathit{\phi,\psi,\omega \in \mathbb{R}^6}$; the virtual torsion and virtual bond angle $\mathit{\kappa,\alpha \in \mathbb{R}^4}$.
\textbf{Node vector features} are position vectors of $\mathit{C_{\alpha}}$, $\mathit{\vec{x_i} \in \mathbb{R}^3}$.
\textbf{Edge scalar features} are established via k-NN (k=16) and the edge length expressed in 32 Radial Basis Functions, $\mathit{e_{RBF} \in \mathbb{R}^{32}}$, as well as the length of the edge itself.
\textbf{Edge vector features} are edge directional unit vectors for both directions $\mathit{\vec{v_{e^{ij}}} = \vec{x_i} - \vec{x_j}}$.
To further prevent overfitting on crystallisation artifacts, random Gaussian noise ($\bar x = 0, \sigma=0.1$\AA) was added to the coordinates \citep{proteinmpnn}.

\subsection{Multi-state GNN} 

DynamicMPNN processes one or multiple protein backbone graphs via a multi-state GNN encoder \citep{grnade}. 
Overall, DynamicMPNN's encoder is equivariant to 3D roto-translation of coordinates as well as ordering of the states in its input.
Encoding is followed by pooling node features across states, which is invariant to the ordering of the states, and autoregressive sequence decoding.

When representing conformational ensembles as a multi-graph, each node feature tensor contains three axes:
(\#nodes, \#conformations, feature channels).
Multi-state GNN's encode multi-graphs by performing message passing on the multi-graph adjacency to \textit{independently} process each conformer, while maintaining permutation equivariance of the updated feature tensors along both the first (\#nodes) and second (\#conformations) axes.

\subsection{Geometric Vector Perceptron layers}
\label{app:model:gvp}

Geometric Vector Perceptrons (GVPs) \citep{jing2021} are a generalization of MLPs to take tuples of scalar and vector features as input and apply $O(3)$-equivariant non-linear updates.
GVP GNN layers process scalar and vector features on separate channels to maintain equivariance. The node scalars $\mathbf{s}_i \in \mathbb{R}^{k \times m}$, node vectors $\mathbf{\vec{v}}_i \in \mathbb{R}^{k \times m' \times 3}$, and edge scalars $\mathbf{e}_{ij}$ and vectors $\mathbf{\vec{e}}_{ij}$ communicate through a message passing operation:
\begin{align}
    \mathbf{m}_{i}, \mathbf{\vec{m}}_{i} &:= \sum_{j \in N_i} \text{GVP} \left( (\mathbf{s}_i, \mathbf{\vec{v}}_i), (\mathbf{s}_j, \mathbf{\vec{v}}_j), \mathbf{e}_{ij}, \mathbf{\vec{e}}_{ij}  \right), \quad\quad &\text{(Message \& aggregate steps)} \\
    \mathbf{s'}_{i}, \mathbf{\vec{v'}}_{i} &:= \text{GVP} \left( (\mathbf{s}_i, \mathbf{\vec{v}}_i), (\mathbf{m}_i, \mathbf{\vec{m}}_i)  \right). \quad\quad &\text{(Update step)}
\end{align}
The overall GNN encoder is $SO(3)$-equivariant due to the use of reflection-sensitive input features (dihedral angles) combined with $O(3)$-equivariant GVP-GNN layers.

\subsection{Conformation order-invariant pooling} 
After using message passing layers that are conformation order-equivariant, 
we add a conformation order-invariant head, which performs average pooling across the conformation channel of the scalar and vector feature tensors, similar to \citet{grnade} (2025): $\mathit{\mathbf{S} \in \mathbb{R}^{n \times k \times m}}$ 
and $\mathit{\mathbf{\vec{V}} \in \mathbb{R}^{n \times k \times m' \times 3}}$ to $\mathit{\mathbf{S} \in \mathbb{R}^{n \times m}}$ and $\mathit{\mathbf{\vec{V}} \in \mathbb{R}^{n \times m' \times 3}}$, 
where $\mathit{n}$ is the sequence length, $\mathit{k}$ is the number of backbones, $\mathit{m}$ is the number of scalar features, and $\mathit{m'}$ is the number of vector features.
The only pooling strategy used in this work is the pooling of the maximum RMSD pair of chains - therefore $\mathit{k=2}$ - although more pooling strategies for homo-oligomers can be used, such as equal averaging of all chains to be inverse folded in the selected PDB entries.





\end{document}

%% file: math_commands.tex
\usepackage{amsmath,amsfonts,bm}

\def\eqref#1{equation~\ref{#1}}

\def\1{\bm{1}}

\def\ve{{\bm{e}}}

\def\vm{{\bm{m}}}

\def\vs{{\bm{s}}}

\def\vv{{\bm{v}}}

\DeclareMathAlphabet{\mathsfit}{\encodingdefault}{\sfdefault}{m}{sl}
\SetMathAlphabet{\mathsfit}{bold}{\encodingdefault}{\sfdefault}{bx}{n}

